\documentclass[10pt,twocolumn,letterpaper]{article}

\usepackage{cvpr}              

\usepackage{multirow}
\usepackage{colortbl}
\usepackage{pifont}
\usepackage[accsupp]{axessibility}

\definecolor{MyDark}{rgb}{0.4,0.4,0.4}
\definecolor{MyGreen}{HTML}{00B050}
\definecolor{MyBlue}{HTML}{058CFA}
\definecolor{cvprblue}{rgb}{0.21,0.49,0.74}
\definecolor{MyGray}{rgb}{0.85, 0.85, 0.85}

\newcommand{\ua}{$\uparrow$}
\newcommand{\da}{$\downarrow$}
\newcommand{\venue}[1]{\footnotesize\textcolor{blue}{[#1]}}

\usepackage{marvosym}

\makeatletter
\renewcommand{\@fnsymbol}[1]{%
    \ifcase#1\or
    \Letter\or
    \ensuremath{\dagger}\or
    \ensuremath{\ddagger}\or
    \ensuremath{\ast}\fi
}
\makeatother

\definecolor{cvprblue}{rgb}{0.21,0.49,0.74}
\usepackage[pagebackref,breaklinks,colorlinks,allcolors=cvprblue]{hyperref}

\def\confName{CVPR}
\def\confYear{2027}

\title{Reconstructing High-Fidelity 3D Part Assemblies via Gaussian Splatting}

\author{Wenhao Li$^{1}$\quad Zixuan Chen$^{1}$\thanks{Corresponding author.}\quad\ \ Jianhuang Lai$^{1,2,3}$\quad Xiaohua Xie$^{1,2,3}$\\
$^1$School of Computer Science and Engineering, Sun Yat-sen University, Guangzhou, China\\
$^2$Guangdong Province Key Laboratory of Information Security Technology, China\\
$^3$Key Laboratory of Machine Intelligence and Advanced Computing, Ministry of Education, China\\
\small\ttfamily{\{liwh258, chenzx3\}@mail2.sysu.edu.cn, \{stsljh, xiexiaoh6\}@mail.sysu.edu.cn}
}

\begin{document}
\maketitle

\begin{abstract}
3D part assembly aims to arrange and connect individual components into functional objects by determining their spatial configurations and assembly relationships.
However, existing methods primarily focus on geometric reasoning and assembly feasibility.
They often rely on predefined CAD models or point clouds and do not directly target high-fidelity 3D assembly reconstruction for immersive visualization and interactive applications.
In this paper, we propose \textbf{AssemblyGS}, the first framework for reconstructing high-fidelity 3D part assemblies from assembly demonstration videos using 3D Gaussian Splatting.
Specifically, we introduce a \textbf{G}aussian \textbf{A}ssembly \textbf{T}ree (GAT) that unifies part-level Gaussian representations, relative assembly poses, and motion parameters within a hierarchical structure.
For each assembly step, we estimate the relative pose between the moving and reference units through occlusion-aware pose tracking, followed by parameterized motion estimation and joint refinement of the assembled pose and motion under a kinematic model.
Finally, we integrate the recovered assembly poses and motions to complete the GAT, with its root representing the complete assembled object.
Experiments on synthetic and real-world data show that \textbf{AssemblyGS} achieves high-quality 3D reconstruction and accurate recovery of assembly poses and motions, enabling 3D part assembly and disassembly.
\end{abstract}

\section{Introduction}\label{sec:intro}
3D part assembly aims to arrange and connect individual components into functional objects by determining their spatial configurations and assembly relationships, playing an important role in robotic manipulation, and virtual manufacturing \cite{wang2021assembly_survey,jiang2022robotic_assembly_survey,ghandi2015assembly_disassembly}.
Existing methods have investigated geometric assembly reconstruction and part configuration estimation from raw scans \cite{functional_assembly}, point clouds \cite{image_part_assembly,rgl_net,category_part_assembly}, and CAD models \cite{joinable}.
Beyond geometric reasoning, other studies have explored instruction-guided assembly understanding and planning \cite{ikea_manual,manual_pa,assemblybench}, as well as video-based assembly understanding from real-world demonstrations \cite{ikea_video_manuals}.

Despite these advances, existing assembly methods still struggle to reconstruct visually realistic 3D assembly assets and support 3D part assembly and disassembly.
Moreover, many methods rely on predefined CAD models or point clouds, whose acquisition and preparation can be labor-intensive when digital models are not readily available.
Together, these limitations highlight the need to reconstruct high-fidelity 3D part assemblies from visual observations.

A natural approach is to reconstruct individual parts from multiview images using 3D Gaussian Splatting \cite{3dgs} and progressively assemble them following the operations observed in an assembly demonstration video.
However, realizing this approach involves three key requirements.
\textbf{First,} a hierarchical representation is needed to organize part-level Gaussians and their assembly relationships for 3D part assembly and disassembly.
\textbf{Second,} severe occlusions and close interactions between moving and reference units make accurate relative pose estimation from videos challenging.
\textbf{Third,} noisy and incomplete pose trajectories make it difficult to recover consistent, parameterized assembly motions for interactive manipulation.
These requirements motivate a unified framework that integrates hierarchical assembly representation with reliable pose and motion estimation.

In this paper, we propose \textit{AssemblyGS}, a novel framework for reconstructing high-fidelity 3D part assemblies from assembly demonstration videos using 3D Gaussian Splatting.
Specifically, we introduce a \textbf{G}aussian \textbf{A}ssembly \textbf{T}ree (GAT) to represent the hierarchical relationships among individual parts and subassemblies, where each assembly operation integrates a moving unit into a reference unit through a relative assembled pose and a parameterized motion.
We first reconstruct individual parts as 3D Gaussians from multi-view images and use their Gaussian representations to initialize the assembly reconstruction.
Given an assembly demonstration video, we process the observed operations sequentially to recover how the individual parts are combined into larger subassemblies.
For each assembly step, we estimate the relative pose of the moving unit \wrt the reference unit through occlusion-aware pose tracking.
We then recover the parameterized motion from the estimated pose trajectory and jointly refine the assembled pose and motion using a kinematic model.
Finally, we complete the GAT by integrating the recovered assembly poses and motions, with its root representing to the final object.

To evaluate \textit{AssemblyGS}, we construct a synthetic assembly dataset, Assembly-20, and conduct real-world experiments to assess reconstruction quality, assembly pose accuracy, and motion estimation performance.
Experiments demonstrate that \textit{AssemblyGS} achieves high-quality 3D reconstruction and accurate recovery of assembly poses and motions, enabling 3D part assembly and disassembly.

Overall, our contributions can be summarized as follows:
\begin{itemize}
    \item To the best of our knowledge, \textit{AssemblyGS} is the first 3DGS-based framework for reconstructing high-fidelity 3D part assemblies from assembly demonstration videos.
    \item We introduce a Gaussian Assembly Tree (GAT) that unifies part-level Gaussian representations, relative assembly poses, and motion parameters within a hierarchical structure.
    By progressively recovering assembly poses and motions from assembly videos, we construct the GAT whose root represents the complete assembled object.
    \item Experiments on synthetic and real-world data demonstrate that \textit{AssemblyGS} achieves high-quality 3D reconstruction and accurate assembly poses and motions, enabling 3D part assembly and disassembly.
\end{itemize}

\section{Related Work}

\noindent\textbf{3D Part Assembly.}
3D part assembly focuses on determining the spatial configurations and connections of individual components to form complete objects, supporting applications in robotic manipulation, and virtual manufacturing \cite{wang2021assembly_survey,jiang2022robotic_assembly_survey,ghandi2015assembly_disassembly}.
Early studies investigated the reconstruction of functional mechanical assemblies from raw 3D scans \cite{functional_assembly}, while PartNet \cite{partnet} introduced a large-scale benchmark for fine-grained and hierarchical 3D part understanding.
Subsequent methods explored image-guided part assembly \cite{image_part_assembly}, progressive part pose estimation through graph reasoning \cite{rgl_net}, and joint prediction between parametric CAD components \cite{joinable}.
Other methods investigated category-level multi-part assembly with joint constraints \cite{category_part_assembly}, instruction-guided assembly understanding and pose estimation \cite{ikea_manual,manual_pa}, and the grounding of assembly instructions in real-world videos \cite{ikea_video_manuals}.
Recently, AssemblyDyno \cite{assemblybench} introduced physics-aware assembly planning for complex industrial objects, while ZAP \cite{zap} explored zero-shot assembly planning using vision-language models.

\noindent\textbf{3D Representations.}
Neural radiance fields (NeRF) \cite{2022nerf} implicitly represent 3D scenes using multilayer perceptrons and synthesize novel views through volume rendering \cite{rendering} and alpha compositing \cite{alpha}.
NeRF-based representations have been extended to sparse-view reconstruction \cite{2023Sparsenerf}, efficient rendering \cite{2022instngp}, generative modeling \cite{2021Giraffe}, medical image reconstruction \cite{cunerf}, and anti-aliasing \cite{2022mip-nerf-360}.
In contrast, 3D Gaussian Splatting (3DGS) \cite{3dgs} explicitly represents scenes using 3D Gaussian primitives and enables efficient optimization and real-time novel-view rendering through differentiable splatting \cite{splatting}.
3DGS has been extended to various applications, including text-to-3D generation \cite{zhu2025segmentdreamer}, avatar synthesis \cite{2024splattingavatar}, single-view reconstruction \cite{szymanowicz2024splatter}, watermarking \cite{guardsplat}, and SLAM \cite{monogs}.
Recent studies \cite{artgs,reartgs,reartgspp,freeartgs} have further explored articulated object reconstruction using 3DGS, enabling part-level geometry reconstruction and articulation parameter estimation.
However, these methods primarily focus on reconstructing articulated objects and modeling their motions across different states, rather than recovering sequential assembly operations to construct complete 3D assemblies from individual parts.

\begin{figure*}[t]
  \centering
  \includegraphics[width=\linewidth]{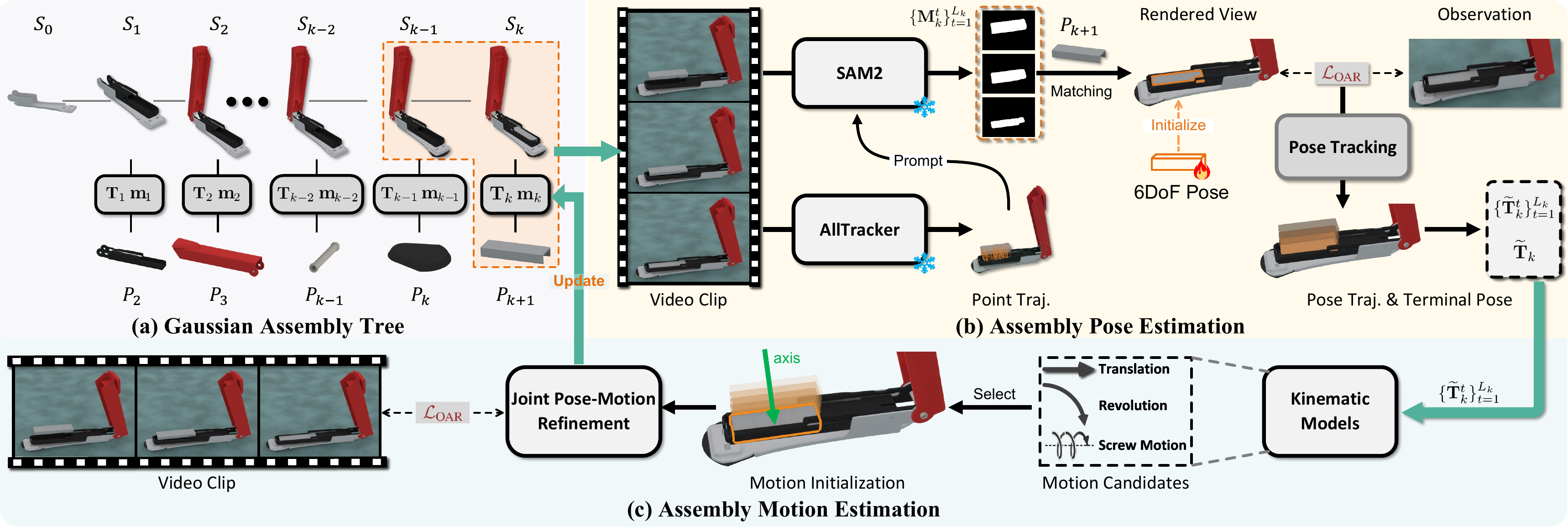}
\caption{\textbf{Overview of our AssemblyGS.}
\textbf{(a)} Gaussian Assembly Tree (GAT) represents the progressive assembly of individual 3D Gaussian parts, where each assembly step incorporates a new part into the existing assembly and records its assembled pose $\mathbf{T}_k$ and motion parameters $\mathbf{m}_k$ on the edge.
\textbf{(b)} Given an assembly video clip, we employ AllTracker \cite{alltracker} and SAM2 \cite{sam2} to identify the moving part and obtain its masks, followed by part matching and occlusion-aware pose tracking based on $\mathcal{L}_\text{OAR}$ in \cref{eq:assembly_render_loss} to estimate the initial 6-DoF pose trajectory and final assembled pose.
\textbf{(c)} Based on the estimated pose trajectory, we initialize candidate kinematic models, including translation, revolute motion, and screw motion, and jointly refine the assembled pose and motion parameters through $\mathcal{L}_\text{OAR}$.
The refined pose and motion parameters are then recorded in GAT to support 3D part assembly and disassembly.
}
  \label{fig:overview}
\end{figure*}

\section{Preliminary: 3D Gaussian Splatting}
\label{sec:preliminary}
3D Gaussian Splatting (3DGS) \cite{3dgs} represents a scene as a collection of anisotropic 3D Gaussians, each parameterized by a mean $\boldsymbol{\mu}\in\mathbb{R}^3$, covariance $\boldsymbol{\Sigma}\in\mathbb{R}^{3\times3}$, opacity $\alpha\in[0,1]$, and view-dependent color $\boldsymbol{c}$ represented by spherical harmonics.
The spatial distribution of each Gaussian is defined as:
\begin{equation}
\mathcal{G}(\mathbf{x};\boldsymbol{\mu},\boldsymbol{\Sigma})=\exp\left(-\frac{1}{2}(\mathbf{x}-\boldsymbol{\mu})^{\top}\boldsymbol{\Sigma}^{-1}(\mathbf{x}-\boldsymbol{\mu})\right).
\label{eq:gaussian}
\end{equation}
Given a camera pose $\mathbf{T}\in\text{SE}(3)$ and intrinsic matrix $\mathbf{K}\in\mathbb{R}^{3\times3}$, each 3D Gaussian is projected onto the image plane through perspective projection, yielding a 2D Gaussian with mean $\hat{\boldsymbol{\mu}}$ and covariance $\hat{\boldsymbol{\Sigma}}$.
The rendered image is then obtained through alpha compositing:
\begin{equation}
\hat{\mathbf{I}}_{x,y}=\sum_{i=1}^{N^\text{g}}\boldsymbol{c}_i\eta_i\prod_{j=1}^{i-1}(1-\eta_j),\quad
\eta_i=\alpha_i\mathcal{G}((x,y);\hat{\boldsymbol{\mu}}_i,\hat{\boldsymbol{\Sigma}}_i),
\label{eq:3dgs_rendering}
\end{equation}
where $N^\text{g}$ denotes the number of Gaussians, and the Gaussians are ordered by depth.
The Gaussian parameters are optimized using an RGB reconstruction loss combining structural similarity and pixel-wise differences.


\section{Method}
In this section, we present \textit{AssemblyGS}, a framework for reconstructing high-fidelity 3D part assemblies from assembly demonstration videos using 3D Gaussian Splatting.
We first introduce the Gaussian Assembly Tree (GAT), which provides a unified hierarchical representation of part geometry, assembly poses, and motion parameters.
For each assembly step, we estimate an initial 6-DoF pose trajectory of the moving part through rendering-based pose tracking.
We then jointly refine the assembled pose and motion parameters under a shared kinematic model.
Finally, we progressively construct the GAT by integrating the recovered assembly poses and motion parameters, with its root representing the fully assembled object, enabling 3D part assembly and disassembly.
The overview is illustrated in \Cref{fig:overview}.

\subsection{Gaussian Assembly Tree}
\label{sec:gat}

Inspired by the assembly tree representation \cite{ikea_manual}, we introduce a Gaussian Assembly Tree (GAT) that integrates 3D Gaussians, relative assembly poses, and motion parameters into a unified hierarchical representation for 3D part assembly reconstruction.
Given an object consisting of $N$ rigid parts, we denote its part set as $\mathcal{P}=\{P_i\}_{i=1}^{N}$.
Each leaf node of GAT represents an individual part and its Gaussian representation, while each internal node represents a subassembly formed by adding a new part to an existing assembly.
The final assembled pose and motion parameters are associated with the incoming part, while the existing assembly serves as the stationary reference.

To initialize GAT, we independently reconstruct the Gaussian representation of each part in its local coordinate frame from calibrated multiview images and corresponding depth maps.
The Gaussian parameters are optimized under RGB and depth supervision to obtain high-fidelity part-level representations.
Throughout our framework, depth maps can be either directly captured or estimated from images using a depth estimation model.

We then progressively construct the hierarchical assembly relationships from an assembly demonstration video consisting of $K$ part-addition steps in chronological order.
Each step $k$ is represented by a video clip $\mathcal{C}_k=\{(\mathbf{I}_k^t,\mathbf{D}_k^t)\}_{t=1}^{L_k}$, containing $L_k$ RGB frames and their corresponding depth maps.
We assume that each step involves a single incoming part with one dominant assembly motion, and index the parts according to their assembly order.
Starting from $\mathcal{S}_0=\{P_1\}$, we progressively update the assembled part set by incorporating a new part at each step:
\begin{equation}
\mathcal{S}_k=\mathcal{S}_{k-1}\cup\{P_{k+1}\}, \quad k=1,\ldots,K.
\label{eq:assembly_tree}
\end{equation}
Each resulting subassembly corresponds to an internal node $v_k$ of GAT, with $\mathcal{S}(v_k)=\mathcal{S}_k$.
For each incoming part, we estimate its final assembled pose $\mathbf{T}_k$ and motion parameters $\mathbf{m}_k$, which are associated with the corresponding GAT connection.
The resulting subassembly is given by
\begin{equation}
\mathcal{H}(\mathcal{S}_k)=\mathcal{H}(\mathcal{S}_{k-1})\cup\mathbf{T}_k\left(\mathcal{H}(P_{k+1})\right),
\label{eq:gat_update}
\end{equation}
where $\mathcal{H}(\cdot)$ denotes the 3D Gaussians of a part or subassembly, and $\mathbf{T}_k$ transforms the incoming part into the coordinate frame of the stationary assembly.
After all $K=N-1$ part-addition steps, $\mathcal{S}_K=\mathcal{P}$ corresponds to the root of GAT, enabling 3D part assembly and disassembly.

\subsection{Assembly Pose Estimation}
\label{sec:assembly_pose}

For each assembly step $k$, we estimate an initial 6-DoF pose trajectory of the moving part relative to the stationary assembly from video clip $\mathcal{C}_k$.
Given the reconstructed 3D Gaussians of the moving part and stationary assembly, together with the reference pose in the camera frame, we recover the moving part poses through rendering-based pose tracking.
All poses are expressed \wrt the stationary assembly and transformed into the camera frame for rendering.

\noindent\textbf{Moving-Part Identification.}
We use AllTracker \cite{alltracker} to track pixels throughout the clip and identify regions of motion.
Specifically, we sample positive prompts from pixels that begin moving first and negative prompts from nearby stationary pixels, and employ SAM2 \cite{sam2} to obtain the moving part masks $\{\mathbf{M}_k^t\}_{t=1}^{L_k}$.
We associate the tracked moving part with the reconstructed parts by matching its observations against a few rendered reference views.

\noindent\textbf{Occlusion-Aware Pose Tracking.}
We first initialize the moving part pose in the first frame through a global search by sampling rotations in $\mathrm{SO}(3)$ and generating translation candidates from the observed depth.
We evaluate these candidates based on their geometric consistency with the observed 3D points, while penalizing poses that occupy observed free space or penetrate the supporting surface.
The highest-scoring candidates are further refined through rendering-based optimization, and the pose with the lowest rendering loss is selected as the initial estimate.
Starting from the initialized pose, we perform occlusion-aware tracking by jointly rendering the moving part and stationary assembly.
Along with the rendered color $\hat{\mathbf{I}}$, depth $\hat{\mathbf{D}}$, and accumulated opacity $\hat{\mathbf{A}}$, we obtain a soft visibility mask $\hat{\mathbf{F}}$ by assigning values of one and zero to the Gaussians of the moving part and stationary assembly, before alpha compositing.
We define the color, mask, and depth losses:
\begin{equation}
\begin{aligned}
\mathcal{L}_{\text{color}} &= \frac{1}{|\mathcal{R}|}\sum_{p\in\mathcal{R}}\left|\hat{\mathbf{I}}(p)-\hat{\mathbf{A}}(p)\mathbf{I}_k^t(p)\right|,\\
\mathcal{L}_{\text{mask}} &= 1-\frac{2\sum_p\hat{\mathbf{F}}(p)\mathbf{M}_k^t(p)}{\sum_p\hat{\mathbf{F}}(p)+\sum_p\mathbf{M}_k^t(p)},\\
\mathcal{L}_{\text{depth}} &= \frac{1}{|\mathcal{D}|}\sum_{p\in\mathcal{D}}\min\!\left(\bigl|\hat{\mathbf{D}}(p)-\hat{\mathbf{A}}(p)\mathbf{D}_k^t(p)\bigr|,c\right),
\end{aligned}
\label{eq:assembly_render_terms}
\end{equation}
where $\mathcal{R}$ denotes the union of the observed moving part mask and the support of the rendered visibility mask, while $\mathcal{D}$ contains pixels within the observed mask with valid depth.
The color and depth losses use opacity-weighted observations, with the depth residual capped at $c$ to reduce the influence of large errors near depth discontinuities.
Occlusion-aware rendering loss is formulated as:
\begin{equation}
\mathcal{L}_{\text{OAR}}=\lambda_{\text{color}}\mathcal{L}_{\text{color}}+\lambda_{\text{mask}}\mathcal{L}_{\text{mask}}+\lambda_{\text{depth}}\mathcal{L}_{\text{depth}}.
\label{eq:assembly_render_loss}
\end{equation}

Starting from the initialized pose, we track the moving part throughout the clip by predicting each subsequent pose under a constant-velocity assumption and refining it by minimizing \cref{eq:assembly_render_loss}.
During tracking, the reconstructed Gaussians and reference pose remain fixed, while only the moving part pose is optimized.
When the moving part becomes largely occluded, we fit a line or an arc to the previously estimated trajectory to guide subsequent pose prediction.
This process yields an initial pose trajectory $\{\widetilde{\mathbf{T}}_k^t\}_{t=1}^{L_k}$ for each assembly step.
We estimate the final assembled pose $\widetilde{\mathbf{T}}_k$ by jointly fitting a shared pose to the final stationary frames of the corresponding clip.
The estimated pose trajectory and final assembled pose serve as initialization for subsequent joint pose and motion refinement.

\subsection{Assembly Motion Estimation}
\label{sec:assembly_motion}

Although rendering-based pose tracking provides an initial pose trajectory, the framewise estimates may drift under occlusion and lack global kinematic consistency.
To address these limitations, we initialize candidate motion models from the tracked trajectory and jointly refine the assembled pose and motion parameters using the corresponding video clip.
We consider three types of assembly motion: translation, revolute motion, and screw motion, assuming a single dominant motion type for each assembly step.

\noindent\textbf{Motion Model Initialization.}
For each assembly step $k$, we first express the tracked poses relative to the estimated assembled pose $\widetilde{\mathbf{T}}_k$ as $\Delta\widetilde{\mathbf{T}}_k^t=\widetilde{\mathbf{T}}_k^t\widetilde{\mathbf{T}}_k^{-1}$.
Based on these relative transformations, we estimate the translation axis from the dominant direction of the part center displacements and initialize the rotation axis by fitting an arc to the center trajectory.
Since the rotation of symmetric components can be difficult to infer from visual observations, we further examine the reconstructed geometry of the moving part and its mating component to identify coaxial threaded structures.
We analyze the radial variation of the Gaussian geometry in cylindrical coordinates $(z,\phi,r)$ around the candidate motion axis, where a thread with signed pitch $h$ exhibits periodicity in the helical phase $(z-h\phi/(2\pi))\bmod |h|$.
We estimate the thread pitch and handedness by evaluating the consistency of these radial variations under candidate helical configurations, and initialize a screw motion when a consistent thread structure is detected.

\noindent\textbf{Joint Pose-Motion Refinement.}
To promote kinematic consistency across frames, we parameterize the moving part pose at frame $t$ of each assembly step as
\begin{equation}
\mathbf{T}_k^t=
\begin{bmatrix}
\mathbf{R}_t & (\mathbf{I}_3-\mathbf{R}_t)\mathbf{o}_k+d_t\mathbf{a}_k\\
\mathbf{0}^{\top} & 1
\end{bmatrix}\mathbf{T}_k,
\label{eq:assembly_pose_refinement}
\end{equation}
where $\mathbf{R}_t=\mathbf{R}(\mathbf{a}_k,\theta_t)$ denotes a rotation of $\theta_t$ around the unit axis $\mathbf{a}_k$, $\mathbf{o}_k$ is a point on the axis, and $d_t$ denotes the axial displacement.
Here, $\mathbf{T}_k$ denotes the final assembled pose, while the motion axis and assembled pose are shared across frames within the assembly step.
To anchor the final assembled configuration, we set $\theta_t=d_t=0$ for the terminal stationary frames, ensuring that $\mathbf{T}_k$ represents the assembled state.
We set $\theta_t=0$ for translation and $d_t=0$ for revolute motion.
For screw motion, we couple rotation and translation through $\theta_t=2\pi d_t/h_k$ while the thread is engaged, where $h_k$ denotes the signed axial advance per revolution.
We introduce an engagement length $\ell_k$ to characterize the axial range of the threaded motion, beyond which the moving part is considered disengaged from assembly.

With the reconstructed Gaussians and reference pose fixed, we jointly refine the assembled pose, motion parameters, and per-frame motion variables over the entire video clip by minimizing the occlusion-aware rendering loss $\mathcal{L}_{\text{OAR}}$ defined in \cref{eq:assembly_render_loss}.
For screw motion, the engagement length $\ell_k$ is additionally optimized as a shared parameter.
We independently optimize the translational and revolute candidates and additionally consider a screw candidate when coaxial threads are detected.
We select the final motion model based on loss, favoring revolute or screw motion only when it sufficiently improves the translational fit.

\noindent\textbf{Assembly Motion Representation.}
We express the selected motion of each assembly step as a twist:
\begin{equation}
\boldsymbol{\xi}_k=
\begin{cases}
(\mathbf{0},\mathbf{a}_k), & \text{translation},\\
(\mathbf{a}_k,\mathbf{o}_k\times\mathbf{a}_k), & \text{revolute motion},\\
(\mathbf{a}_k,\mathbf{o}_k\times\mathbf{a}_k+\frac{h_k}{2\pi}\mathbf{a}_k), & \text{screw motion}.
\end{cases}
\label{eq:assembly_twist}
\end{equation}
Using the refined assembled pose $\mathbf{T}_k$ and motion twist $\boldsymbol{\xi}_k$, we represent the continuous motion of the incoming part as
\begin{equation}
\mathbf{T}_k(s)=\exp(s\hat{\boldsymbol{\xi}}_k)\mathbf{T}_k,\quad s\in[0,s_k],
\label{eq:assembly_motion_path}
\end{equation}
where $\hat{\boldsymbol{\xi}}_k$ denotes the matrix representation of the twist, $s=0$ corresponds to the assembled configuration, and $s_k$ denotes the recovered motion extent.
The motion direction is chosen to reverse the demonstrated assembly operation, with $s$ representing axial displacement for translation and rotation angle for revolute and screw motion.
For screw motion, the recovered motion is restricted to the threaded engagement range.
The selected motion type, axis parameters, and motion extent, together with the engagement length for screw motion, constitute the motion parameters $\mathbf{m}_k$.
The refined assembled pose $\mathbf{T}_k$ and motion parameters $\mathbf{m}_k$ are associated with the corresponding GAT connection.

\subsection{3D Part Assembly and Disassembly}
\label{sec:assembly_disassembly}

\noindent\textbf{3D Part Assembly.}
Starting from the reconstructed Gaussian parts, we progressively assemble the object by incorporating new parts in the demonstrated assembly order.
At each step, the incoming part is transformed into its assembled configuration using the estimated pose $\mathbf{T}_k$, while the corresponding motion parameters $\mathbf{m}_k$ are retained for subsequent disassembly.
After all assembly steps, the resulting 3D assembly, together with the recorded assembly poses and motion parameters, defines the complete GAT.

\noindent\textbf{3D Part Disassembly.}
Given the completed GAT, we disassemble the object by reversing the recorded assembly pose and motion parameters.
At each step, we keep the remaining assembly fixed and move the most recently added part along the recovered motion path $\mathbf{T}_k(s)$ from $s=0$ to $s=s_k$.
We repeat this process until all added parts have been removed, recovering the individual Gaussian parts.

\section{Experiments}
\subsection{Experimental Setups}
\noindent\textbf{Datasets.} 
We evaluate the effectiveness of our proposed \textit{AssemblyGS} on the synthetic and real-world datasets.

\noindent\textbf{New Benchmark: Assembly-20.}
As there is no existing benchmark specifically designed for 3D part assembly reconstruction, we propose Assembly-20, a new synthetic benchmark containing 20 assemblies from the Fusion 360 Gallery Assembly Dataset \cite{joinable}, spanning mechanical, industrial, and everyday objects.
Each assembly consists of 5 to 15 parts, with a total of 132 parts across the dataset.
For each assembly, we provide RGB-D demonstration videos captured from a fixed camera, along with the corresponding 3D Gaussian representations of individual parts.
The benchmark contains 104 translational, 25 screw, and 3 revolute motions associated with standard assembly operations, together with 3 additional translational motions involving post-assembly interactions.
The additional motions account for operations beyond part installation, such as pressing assembled components.
Each operation is annotated with ground-truth assembled poses and motion parameters.

\noindent\textbf{Real-world Assemblies.}
We collect four real-world assemblies, including an earbud case, an event camera, a remote control, and a toy train.
These objects involve diverse assembly operations, including battery insertion, lid rotation, and cover sliding, covering translational, revolute, and screw motions.
For each object, we capture calibrated multiview images of its individual parts for 3DGS reconstruction and record assembly demonstration videos using a fixed Intel RealSense D435i camera.
Depth maps are estimated from infrared stereo pairs using FoundationStereo \cite{foundationstereo} and aligned with the corresponding RGB images.
Camera poses are calibrated using a ChArUco board.
The videos are divided into clips corresponding to individual motion stages.
These data enable qualitative evaluation of 3D assembly reconstruction and part-level motion recovery under real-world imaging conditions and occlusions.

\noindent\textbf{Baselines.}
Since existing methods do not directly address 3D part assembly reconstruction from demonstration videos, we compare \textit{AssemblyGS} with two closely related assembly methods: RPF \cite{rpf} and AssemblyDyno \cite{assemblybench}.
To adapt these methods to our benchmark, we provide the point clouds of individual parts and the assembly order.
RPF employs flow matching to generate assembled point clouds and recover the corresponding part poses.
AssemblyDyno predicts assembled poses and assembly trajectories from step-by-step instruction manuals, which we generate using its provided templates and ground-truth assembly information.
We additionally replace our pose tracking procedure with FoundationPose \cite{foundationpose}, while retaining the remaining components of \textit{AssemblyGS}.
All methods are evaluated using their native geometric or motion outputs without introducing additional appearance reconstruction modules.

\noindent\textbf{Evaluation Metrics.}
We evaluate \textit{AssemblyGS} in terms of assembly pose accuracy, assembly motion accuracy, and reconstruction quality.
For consistent geometric evaluation, distance thresholds and errors are defined relative to the diameter $d$ of each part, measured as the maximum distance between any two points on its surface \cite{hinterstoisser2012model}.

\noindent\textbf{\textit{1) Assembled Pose Accuracy.}}
We evaluate assembly pose accuracy using ADD-S \cite{hinterstoisser2012model}, Part Accuracy (PA) \cite{image_part_assembly}, and Success Rate (SR) \cite{li2023rearrangement}.
ADD-S measures part-level pose errors while accounting for geometric symmetries.
A part is considered correctly assembled if its ADD-S error is below $5\%$ of its diameter.
PA measures the percentage of correctly assembled parts, while SR measures the percentage of assemblies in which all parts are correctly assembled.

\noindent\textbf{\textit{2) Assembly Motion Accuracy.}}
We evaluate assembly motion estimation using motion type accuracy (Type Acc.), axis angular error (Axis Ang.), axis position error (Axis Pos.), and part motion error (Motion Err.).
Type Acc. measures the percentage of correctly classified motion types, including translation, revolute, and screw motions.
Axis Ang. measures the angular deviation between predicted and ground-truth motion directions, while Axis Pos. measures the positional discrepancy between motion axes for revolute and screw motions.
Motion Err. reports two metrics: geodesic rotation error ($^\circ$) and Euclidean translation error (cm), presented in this order.

\noindent\textbf{\textit{3) Reconstruction Quality.}}
We evaluate the novel-view synthesis quality of reconstructed 3D assemblies at each assembly stage using PSNR, SSIM \cite{ssim}, and LPIPS \cite{lpips}.
These metrics measure photometric accuracy, structural similarity, and perceptual quality between rendered and ground-truth views, respectively.

\begin{table*}[!t]
\caption{
  \textbf{Quantitative comparison on Assembly-20.}
  We evaluate assembly pose accuracy, assembly motion accuracy, and reconstruction quality.
  Motion error is computed on trajectories expressed \wrt the terminal pose of each clip.
  \textbf{Bold} value indicates the best performance.
}
\label{tab:quantitative_comp}
\centering
\resizebox{\textwidth}{!}{
\begin{tabular}{lccccccccccccc}
\toprule

\multirow{2}{*}{Method} &&\multicolumn{3}{c}{Assembly Pose}&&\multicolumn{4}{c}{Assembly Motion}&&\multicolumn{3}{c}{Reconstruction Quality} \\
\cmidrule{3-5}\cmidrule{7-10}\cmidrule{12-14}
&&ADD-S (mm)\da & PA (\%)\ua & SR (\%)\ua && Type Acc. (\%)\ua & Axis Ang. ($^\circ$)\da & Axis Pos. (\%d)\da & Motion Err. ($^\circ$ / cm)\da &&PSNR\ua & SSIM\ua & LPIPS\da \\
\midrule
RPF \cite{rpf} \venue{NeurIPS'25} &
&3.247 &57.65 &20 &
&-- &-- &-- &-- &
&-- &-- &-- \\
AssemblyDyno \cite{assemblybench} \venue{CVPR'26} &
&11.166 &20.51 &5 &
&77.27 &-- &-- &20.853 / 3.331 &
&-- &-- &-- \\
\midrule
\textbf{AssemblyGS (Ours)} + FoundationPose \cite{foundationpose} &
&6.987 &88.90 &50 &
&97.04 &6.100 &2.030 &17.847 / 1.435 &
&28.13 &0.9701 &0.03408 \\
\rowcolor{gray!15}
\textbf{AssemblyGS (Ours)} &
&\textbf{0.812} & \textbf{97.04} & \textbf{90} &&\textbf{99.26} & \textbf{0.197} & \textbf{0.760} & \textbf{1.841} / \textbf{0.172} &&\textbf{30.07} & \textbf{0.9774} & \textbf{0.02724}\\
\bottomrule
\end{tabular}
}
\end{table*}

\begin{figure*}[t]
  \centering
  \includegraphics[width=\linewidth]{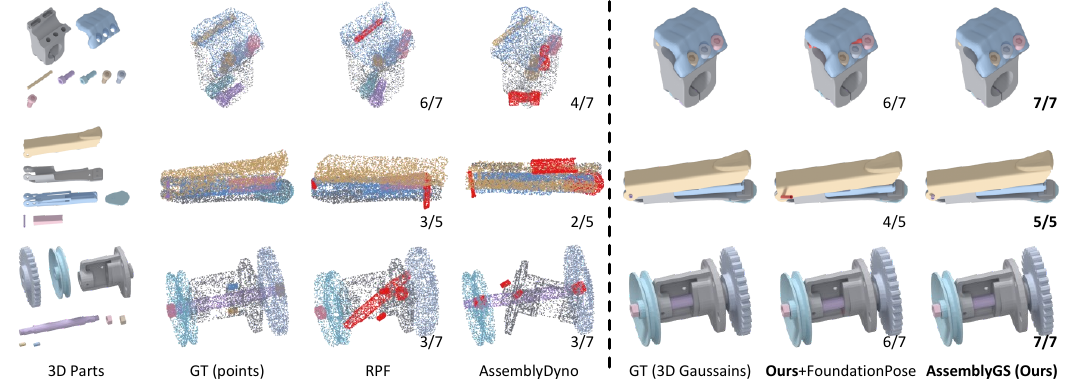}
  \caption{\textbf{Visual comparison on Assembly-20.}
  The leftmost column shows the individual 3D Gaussian parts.
  RPF \cite{rpf} and AssemblyDyno \cite{assemblybench} are visualized as point clouds, alongside the corresponding ground-truth point clouds.
  The results of our variant using FoundationPose \cite{foundationpose} and our full \textit{AssemblyGS} are rendered using 3DGS, alongside the corresponding ground-truth 3D Gaussians.
  The numbers at the bottom-right corner of each method's result denote the number of correctly assembled parts \wrt the total number of parts.
}
  \label{fig:vis_comp}
\end{figure*}


\noindent\textbf{Implementation Details.}
\textit{AssemblyGS} is built in Splatfacto based on Nerfstudio \cite{nerfstudio} and GSplat \cite{ye2025gsplat}.
All experiments are conducted on a single NVIDIA RTX 3090 GPU.

\noindent\textit{\textbf{1) Gaussian Part Reconstruction.}}
We independently reconstruct each part with RGB and depth supervision.
Each part is optimized for 30000 iterations, with the depth loss weight set to $1.0$.
The densification gradient threshold is set to $2\times10^{-4}$, which is stopped after 25000 iterations.

\noindent\textit{\textbf{2) Assembly Pose Estimation.}}
We employ pretrained AllTracker \cite{alltracker} and SAM2.1 \cite{sam2} models to obtain moving-part masks.
For pose initialization, we sample 4800 rotations by combining 200 uniformly distributed directions with 24 in-plane rotations.
We retain the eight best candidates for rendering-based refinement, each optimized for up to 150 iterations.
During pose tracking, we use Adam \cite{adam} with a learning rate of $10^{-2}$ and a maximum of 60 iterations per frame, with early stopping.
The $\mathcal{L}_\text{OAR}$ weights are set to $\lambda_{\mathrm{color}}=1$, $\lambda_{\mathrm{mask}}=1$, and $\lambda_{\mathrm{depth}}=0.5$.
The terminal pose of each motion stage is jointly estimated from the final five stationary frames of the corresponding clip.

\noindent\textit{\textbf{3) Assembly Motion Estimation.}}
We refine each candidate motion using Adam for up to 300 iterations, while keeping the Gaussian representations and reference pose fixed.
For each motion stage, we uniformly sample up to 80 moving frames and include the final five stationary frames.
The learning rates are set to $5\times10^{-3}$ for the shared motion parameters and terminal pose, and $2\times10^{-3}$ for the per-frame motion variables, with a 20-iteration warm-up.

\subsection{Experimental Results}
We first conduct quantitative and visual comparisons between our \textit{AssemblyGS} and state-of-the-art methods on the synthetic Assembly-20 benchmark.
We then demonstrate 3D part assembly and disassembly, as well as part-level motion, on both synthetic and real-world data.

\noindent\textbf{Quantitative Comparisons.}
\Cref{tab:quantitative_comp} presents quantitative comparisons between \textit{AssemblyGS} and competing methods on the synthetic Assembly-20 benchmark in terms of assembly pose accuracy, assembly motion accuracy, and reconstruction quality.
As shown, \textit{AssemblyGS} achieves the best performance across all evaluated metrics.
For assembly pose estimation, our method achieves an ADD-S of $0.812$, a PA of $97.04\%$, and an SR of $90\%$, substantially outperforming existing methods.
For assembly motion estimation, \textit{AssemblyGS} achieves a motion type accuracy of $99.26\%$ and significantly reduces axis angular and positional errors to $0.197^\circ$ and $0.760$, respectively.
In terms of Motion Err., our method reduces the geodesic rotation and Euclidean translation errors to $1.841^\circ$ and $0.172$ cm, respectively, demonstrating its ability to accurately recover assembly motions.
Furthermore, compared with the variant using FoundationPose \cite{foundationpose}, \textit{AssemblyGS} improves PSNR by $1.94$ dB, while achieving higher SSIM and lower LPIPS, demonstrating its effectiveness in reconstructing high-fidelity 3D assemblies.

\begin{table*}[!t]
\caption{\textbf{Ablation study on the synthetic Assembly-20 benchmark.}
We evaluate the effectiveness of the proposed occlusion-aware pose tracking (OAPT) and joint pose-motion refinement (JPMR) modules.
Motion error is computed on trajectories expressed \wrt the terminal pose of each clip.
\textbf{Bold} values indicate the best performance.}
\label{tab:ablation}
\centering
\resizebox{\textwidth}{!}{
\begin{tabular}{ccccccccccccccc}
\toprule
\multirow{2}{*}{OAPT}&\multirow{2}{*}{JPMR}&&\multicolumn{3}{c}{Assembly Pose}&&\multicolumn{4}{c}{Assembly Motion}&&\multicolumn{3}{c}{Reconstruction Quality} \\
\cmidrule{4-6}\cmidrule{8-11}\cmidrule{13-15}
         &         &&ADD-S (mm)\da & PA (\%)\ua & SR (\%)\ua && Type Acc. (\%)\ua & Axis Ang. ($^\circ$)\da & Axis Pos. (\%d)\da & Motion Err. ($^\circ$ / cm)\da &&PSNR\ua & SSIM\ua & LPIPS\da \\
\midrule
         &         &&5.503 & 94.07 & 80 &&97.78 & 0.857 & 2.401 & 12.216 / 1.122 &&30.03 & 0.9771 & 0.02738 \\
\ding{52}&         &&0.893 & 94.07 & 80 &&97.78 & 0.417 & 2.012 & 12.625 / 0.739 &&30.03 & 0.9772 & 0.02736 \\
\rowcolor{gray!15}
\ding{52}&\ding{52}&&\textbf{0.812} & \textbf{97.04} & \textbf{90} &&\textbf{99.26} & \textbf{0.197} & \textbf{0.760} & \textbf{1.841} / \textbf{0.172} &&\textbf{30.07} & \textbf{0.9774} & \textbf{0.02724}\\
\bottomrule
\end{tabular}
}
\end{table*}

\begin{figure*}[t]
  \centering
  \includegraphics[width=\linewidth]{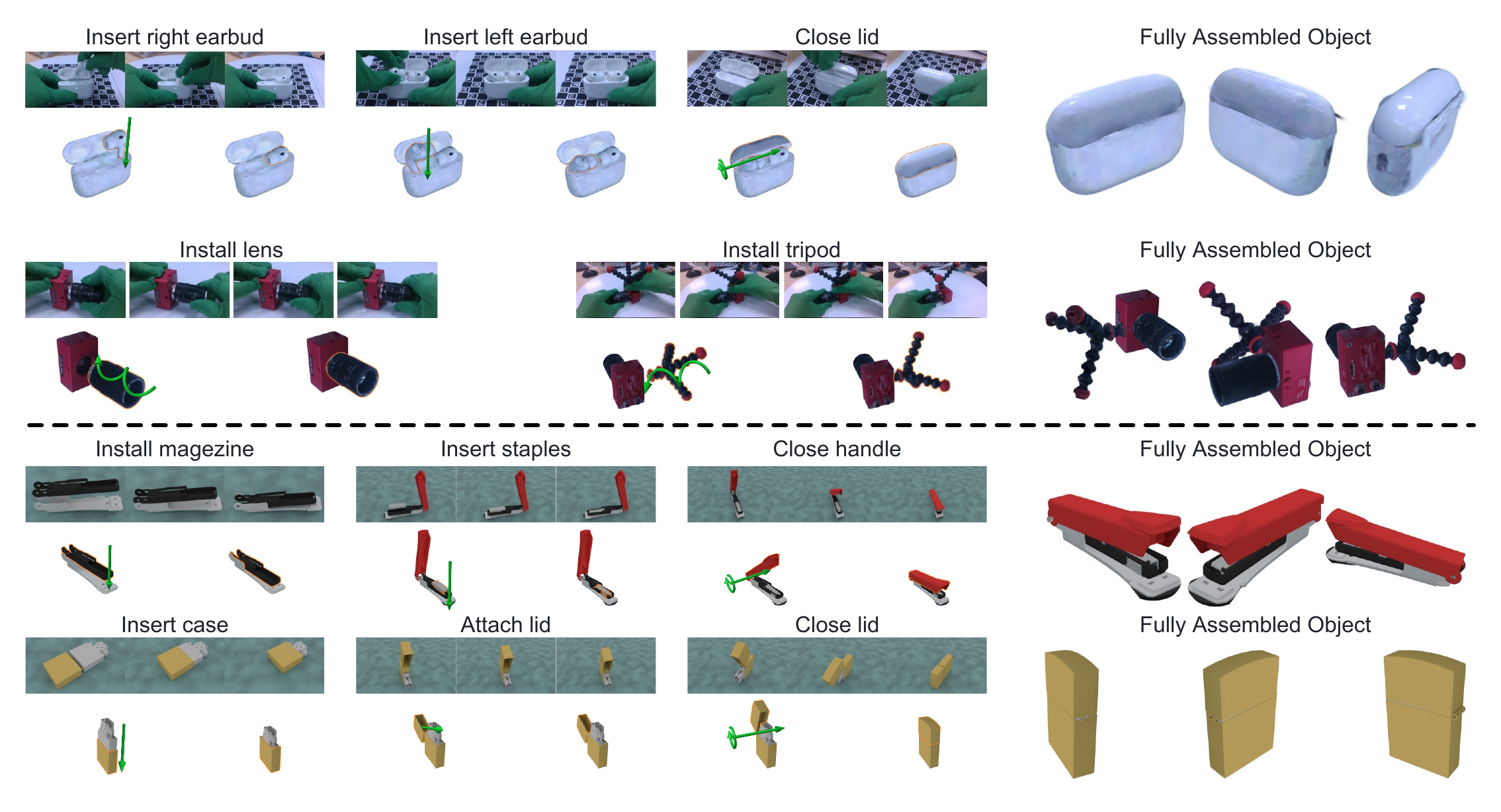}
  \caption{\textbf{Visualization of 3D part assembly reconstruction.}
  We present detailed assembly processes on real-world assemblies (top two rows) and the synthetic Assembly-20 benchmark (bottom two rows).
  For each assembly step, the upper images show the input assembly demonstration video, while the lower images visualize the estimated part poses, assembly motions, and intermediate 3D Gaussian reconstructions.
  The green arrows indicate the recovered assembly motions.
  The rightmost column presents novel-view renderings of the fully assembled objects, demonstrating the ability of \textit{AssemblyGS} to reconstruct high-fidelity 3D assemblies from visual demonstrations.
}
  \label{fig:3d_part_assembly_vis}
\end{figure*}

\noindent\textbf{Visual Comparisons.}
\Cref{fig:vis_comp} presents visual comparisons of the fully assembled objects on representative scenes from Assembly-20.
For RPF \cite{rpf} and AssemblyDyno \cite{assemblybench}, we visualize their assembly results using point clouds to facilitate comparison.
As shown, both methods exhibit noticeable assembly errors, with several parts incorrectly positioned or oriented.
Although our variant incorporating FoundationPose \cite{foundationpose} produces visually realistic reconstructions, it still suffers from inaccurate part placement in some cases.
In contrast, \textit{AssemblyGS} accurately recovers the assembled configurations of all parts in these examples, producing high-fidelity 3D assemblies with accurate part placement.
These results further demonstrate the effectiveness of our method in recovering accurate assembly poses while maintaining high reconstruction quality.

\noindent\textbf{3D Part Assembly Reconstruction.}
\Cref{fig:3d_part_assembly_vis} illustrates the progressive assembly reconstruction processes of \textit{AssemblyGS} on both synthetic and real-world data.
For each assembly step, we visualize the estimated part poses, assembly motions, and intermediate 3D Gaussian reconstructions alongside the corresponding demonstration video frames.
As shown, \textit{AssemblyGS} successfully recovers the assembly motions and progressively reconstructs the complete objects by accurately positioning individual parts.
The novel-view renderings of fully assembled objects demonstrate the ability of our method to produce high-fidelity 3D assemblies.


\subsection{Ablation Study}
We conduct ablation studies to evaluate the contributions of the occlusion-aware pose tracking (OAPT) and joint pose-motion refinement (JPMR) modules.
All experiments are conducted on the synthetic Assembly-20 benchmark.
As shown in \Cref{tab:ablation}, we progressively incorporate OAPT and JPMR into a basic rendering-based pose tracking framework (row 1).
Introducing OAPT (row 2) reduces ADD-S from $5.503$ to $0.893$ and decreases axis angular error by $51.3\%$, demonstrating its effectiveness in improving pose estimation under occlusion.
Building upon OAPT, JPMR (row 3) further increases PA from $94.07\%$ to $97.04\%$ and SR from $80\%$ to $90\%$, while reducing axis angular and positional errors by $52.8\%$ and $62.2\%$, respectively.
In terms of Motion Err., JPMR also substantially reduces the geodesic rotation and Euclidean translation errors from $12.625^\circ$ and $0.739$ cm to $1.841^\circ$ and $0.172$ cm, respectively, demonstrating the effectiveness of joint pose-motion refinement under a shared kinematic model.
Overall, \textit{AssemblyGS} reduces ADD-S by $85.2\%$ compared with the baseline, achieving substantial improvements in assembly pose and motion accuracy while preserving high reconstruction quality.

\section{Conclusion}
In this paper, we present \textit{AssemblyGS}, the first 3DGS-based framework for reconstructing high-fidelity 3D part assemblies from assembly demonstration videos.
We introduce a Gaussian Assembly Tree (GAT) that integrates part-level 3D Gaussians, relative assembly poses, and motion parameters into a unified structure.
By progressively estimating assembly poses and parameterized motions from video clips, \textit{AssemblyGS} reconstructs complete 3D assemblies and enables 3D part assembly and disassembly.
Experiments on Assembly-20 and real-world data show our effectiveness in high-quality 3D reconstruction and accurate assembly pose and motion estimation.
Overall, our work provides a foundation for high-fidelity 3D assembly creations from videos.

\section*{Acknowledgments}
This work was supported in part by the National Natural Science Foundation of China (Grant No. U22A2095) and in part by the Key Areas Research Plan of the Guangdong S\&T Programme (Grant No. 2025B0101120008).

\section*{LLM Usage Statement}
ChatGPT (OpenAI) was used solely for language polishing, grammatical correction, and improving the clarity of the manuscript.
The authors independently verified all scientific content and experimental results, reviewed and approved the final submitted version, and take full responsibility for all submitted content.

{
    \small
    \bibliographystyle{ieeenat_fullname}
    \bibliography{main}
}

\clearpage
\renewcommand\thesection{\Alph{section}}
\renewcommand\thefigure{S\arabic{figure}}
\renewcommand{\thetable}{S\arabic{table}}
\setcounter{page}{1}
\setcounter{section}{0}
\setcounter{figure}{0}
\setcounter{table}{0}

\maketitlesupplementary

\section{Overview}
This supplementary material provides additional visual results to further demonstrate the effectiveness of \textit{AssemblyGS}, organized as follows:
\begin{itemize}
\item Section \ref{sec:visual_part_level_motion} presents additional visualizations of part-level motions, demonstrating the recovered motion parameters and articulated configurations of 3D assemblies.
\item Section \ref{sec:visual_comp_finals} presents additional visual comparisons of fully assembled objects reconstructed by different methods on the synthetic Assembly-20 benchmark.
\item Section \ref{sec:visual_assembly_steps} provides detailed visualizations of progressive 3D part assembly reconstruction, including estimated assembly poses, motions, and intermediate reconstruction results on synthetic and real-world data.
\end{itemize}

\section{Part-Level Motion}
\label{sec:visual_part_level_motion}
\begin{figure}
    \includegraphics[width=\linewidth]{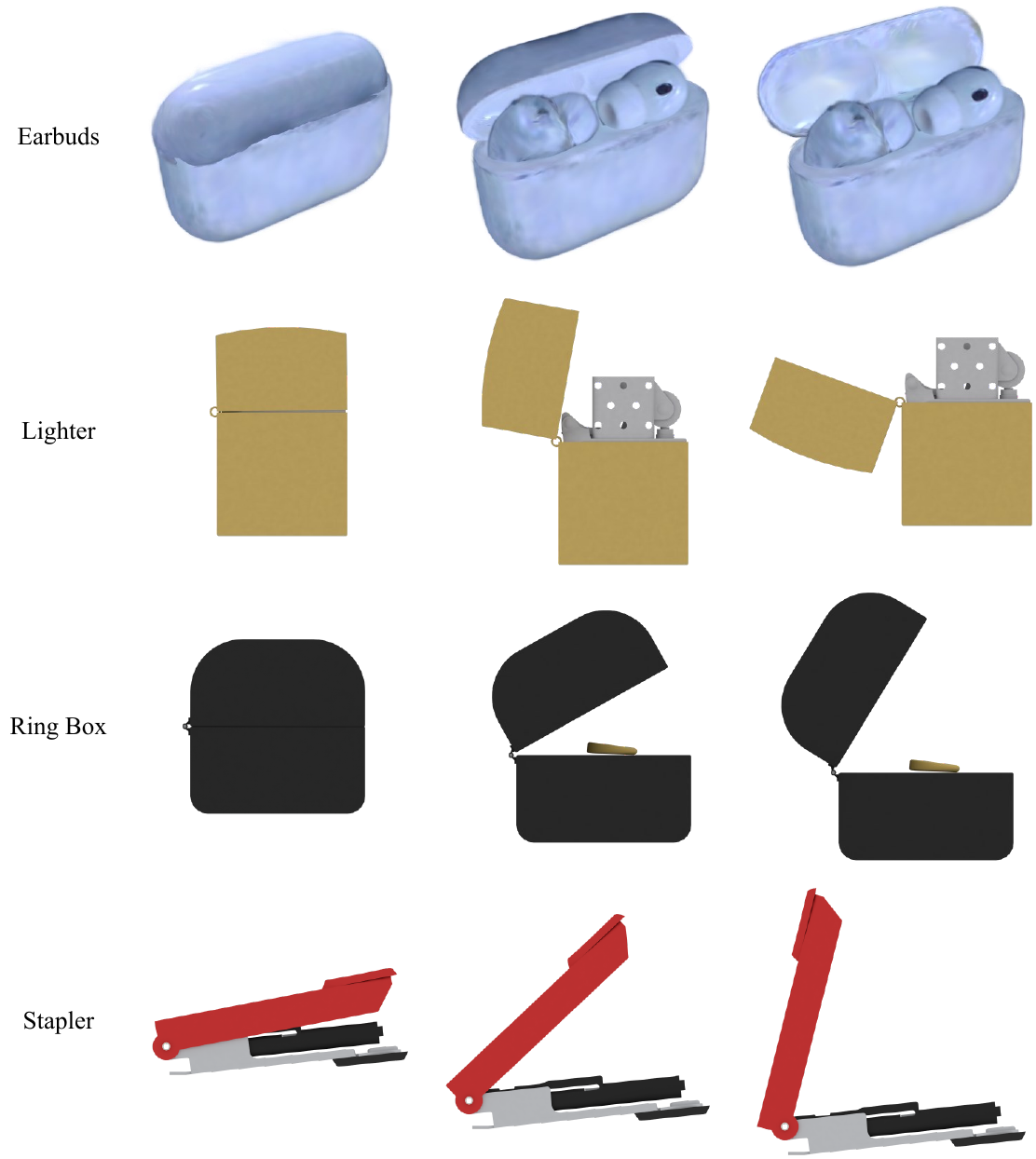}
    \caption{\textbf{Visualization of part-level motions after assembly.}
    We present results on a real-world earbud case (first row) and three synthetic objects from Assembly-20 (remaining rows), including a lighter, a ring box, and a stapler.
    Each row illustrates different configurations of a fully assembled object, showing that individual parts can still move according to their recovered motion parameters after assembly.
    These results demonstrate that \textit{AssemblyGS} preserves part-level motion capabilities while maintaining high-fidelity 3D reconstruction quality across different configurations.
    }
    \label{fig:part_level_motion}
\end{figure}

We further demonstrate that the 3D assemblies reconstructed by \textit{AssemblyGS} retain the motion capabilities of individual parts after assembly.
As shown in \Cref{fig:part_level_motion}, the fully assembled objects can exhibit different configurations by moving individual parts according to their recovered motion parameters.
These results demonstrate that \textit{AssemblyGS} not only reconstructs high-fidelity 3D assemblies but also preserves their part-level motion capabilities, enabling the visualization of different articulated configurations after assembly.

\section{Additional Visual Comparisons}
\label{sec:visual_comp_finals}
\begin{figure*}
    \includegraphics[width=\textwidth]{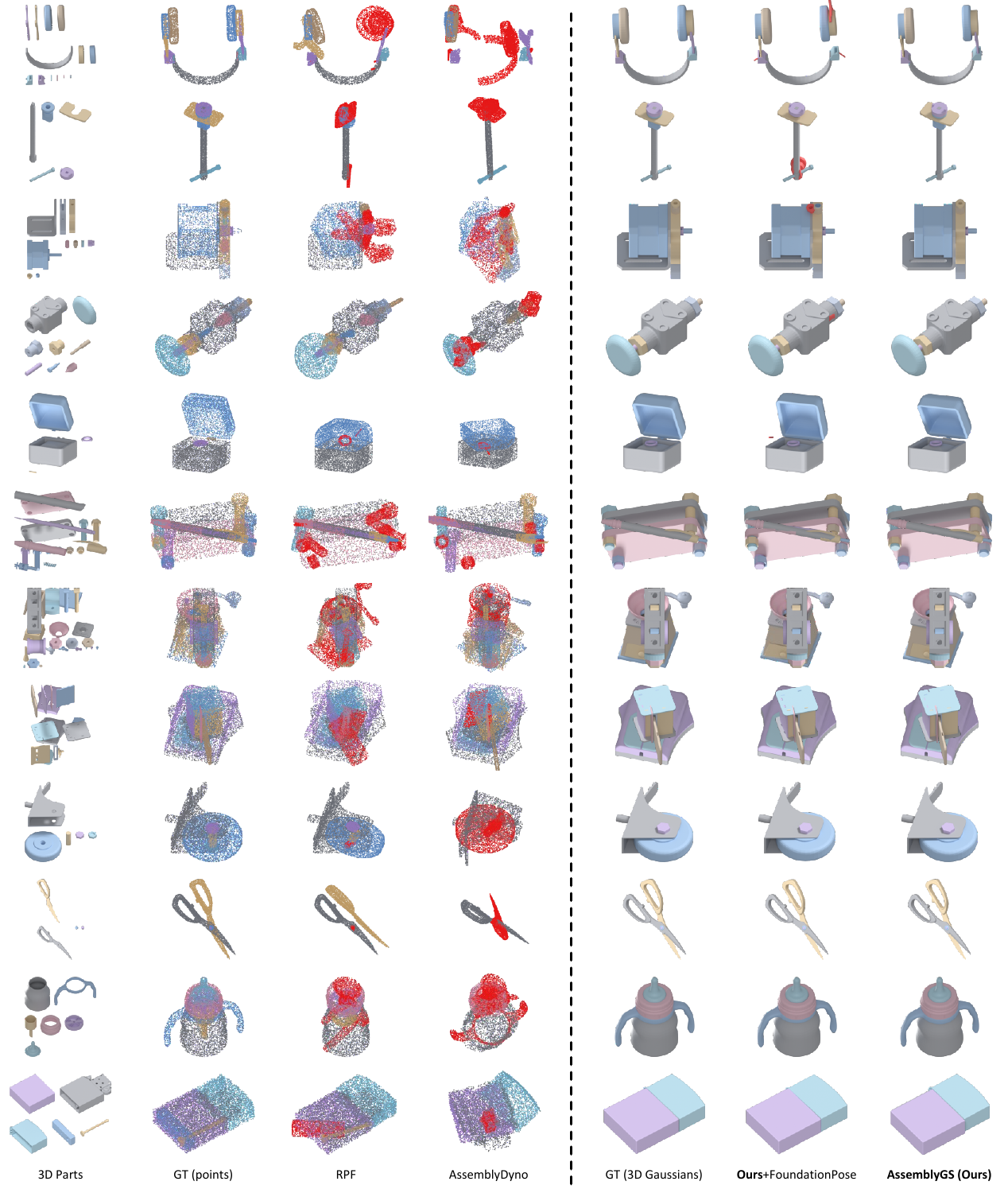}
    \caption{\textbf{Additional visual comparison on Assembly-20.}
    The leftmost column shows the individual 3D Gaussian parts.
    RPF \cite{rpf} and AssemblyDyno \cite{assemblybench} are visualized as point clouds, alongside the corresponding ground-truth point clouds.
    The results of our variant using FoundationPose \cite{foundationpose} and our full \textit{AssemblyGS} are rendered using 3DGS, alongside the corresponding ground-truth 3D Gaussians.
    }
    \label{fig:visual_comp_finals}
\end{figure*}

We provide additional visual comparisons between \textit{AssemblyGS} and competing methods on the synthetic Assembly-20 benchmark.
As shown in \Cref{fig:visual_comp_finals}, the results of RPF \cite{rpf} and AssemblyDyno \cite{assemblybench} are visualized as point clouds, while our variant using FoundationPose \cite{foundationpose} and the full \textit{AssemblyGS} are rendered using 3DGS.
Compared with existing methods, \textit{AssemblyGS} achieves more accurate part placement and consistently reconstructs high-fidelity 3D assemblies across diverse objects.
These additional results further demonstrate the effectiveness of our method in recovering accurate assembly configurations while preserving high reconstruction quality.

\begin{figure*}
    \includegraphics[width=\textwidth]{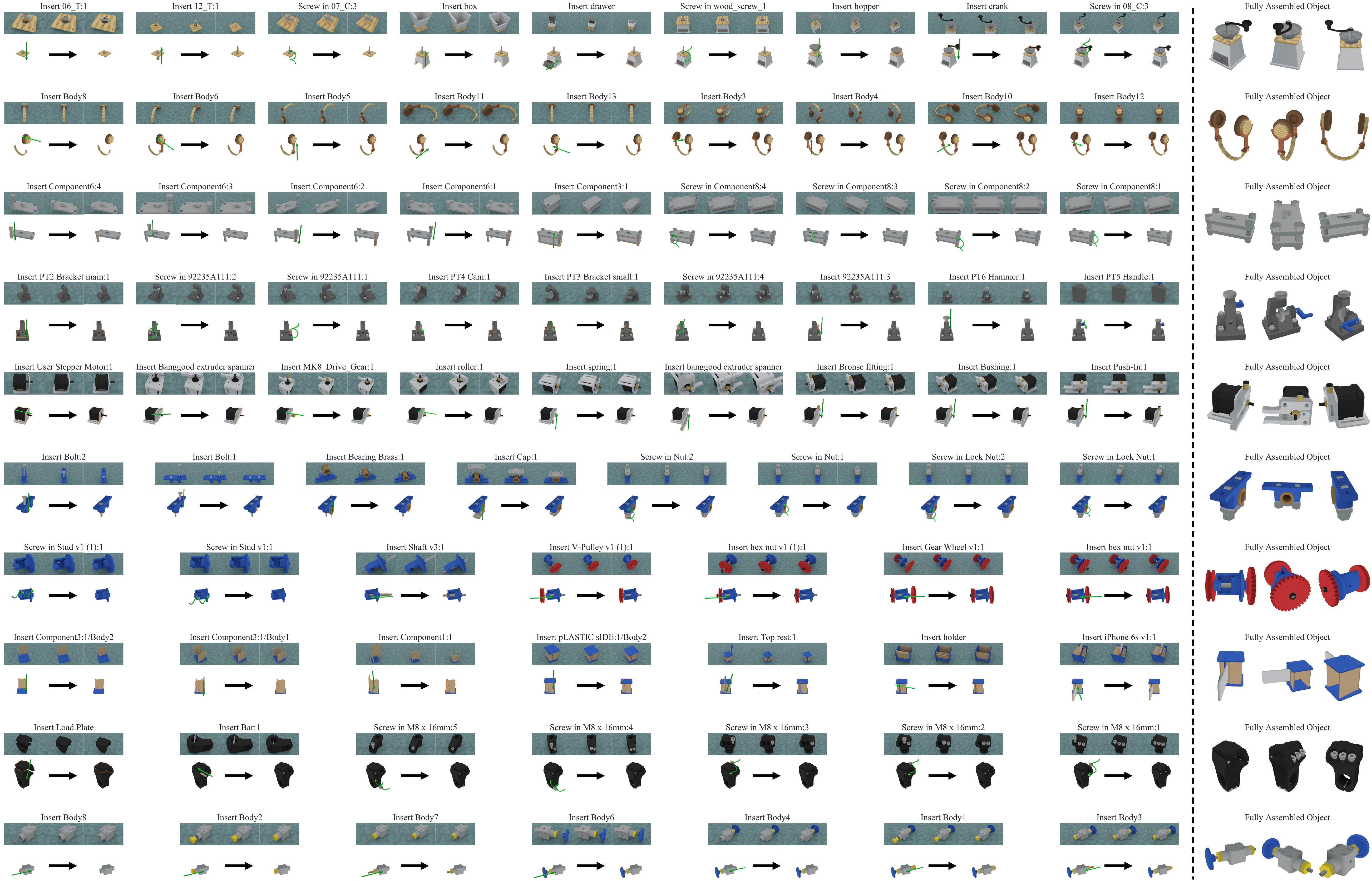}
    \caption{\textbf{Additional visualizations of 3D part assembly reconstruction with more assembly steps.}
    We present detailed assembly reconstruction processes for objects with relatively large numbers of assembly steps from the synthetic Assembly-20 benchmark.
    For each assembly step, the upper images show the input assembly demonstration video frames, while the lower images visualize the estimated part poses, assembly motions, and intermediate 3D Gaussian reconstructions.
    The green arrows indicate the recovered assembly motions.
    The rightmost column presents novel-view renderings of the fully assembled objects.
    For clarity, some intermediate assembly steps are omitted in the first two rows.
    }
    \label{fig:assembly_steps_syn_long}
\end{figure*}

\begin{figure*}
    \includegraphics[width=\textwidth]{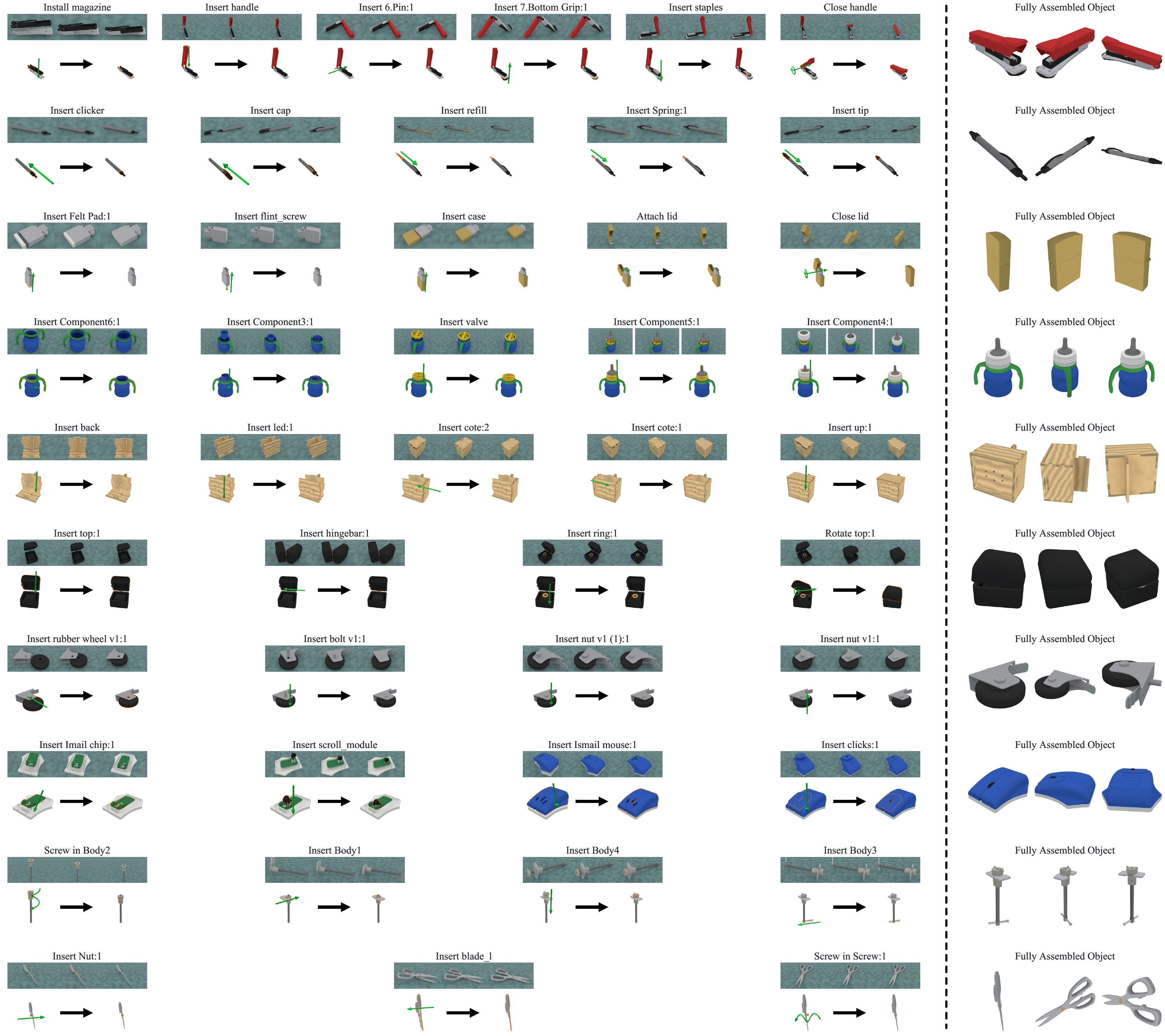}
    \caption{\textbf{Additional visualizations of 3D part assembly reconstruction with fewer assembly steps.}
    We present detailed assembly reconstruction processes for objects with relatively few assembly steps from the synthetic Assembly-20 benchmark.
    For each assembly step, the upper images show the input assembly demonstration video frames, while the lower images visualize the estimated part poses, assembly motions, and intermediate 3D Gaussian reconstructions.
    The green arrows indicate the recovered assembly motions.
    The rightmost column presents novel-view renderings of the fully assembled objects.
    }
    \label{fig:assembly_steps_syn_short}
\end{figure*}

\begin{figure*}
    \includegraphics[width=\textwidth]{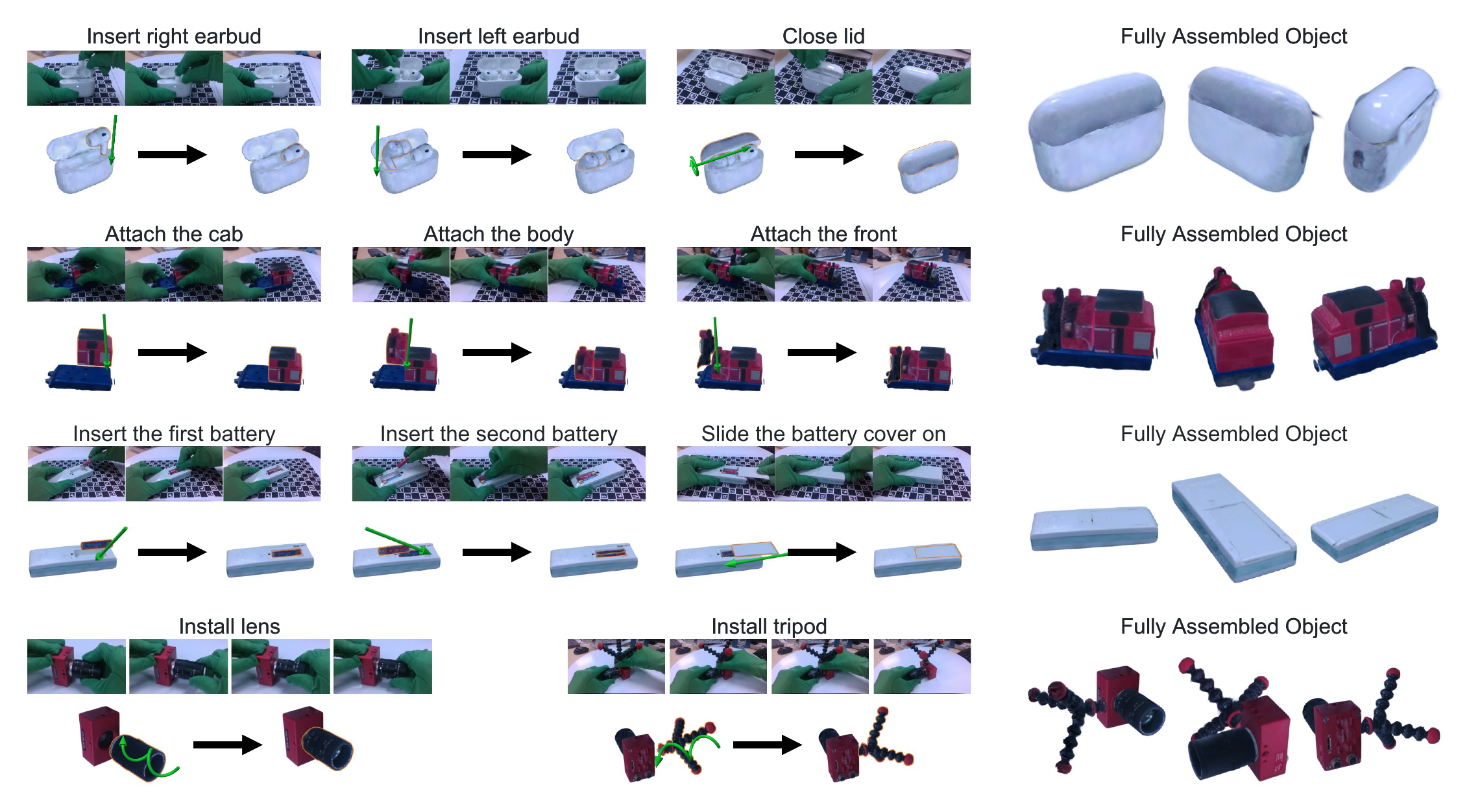}
    \caption{\textbf{Additional visualizations of 3D part assembly reconstruction on real-world data.}
    We present detailed assembly reconstruction processes for real-world 3D assemblies.
    For each assembly step, the upper images show the input assembly demonstration video frames, while the lower images visualize the estimated part poses, assembly motions, and intermediate 3D Gaussian reconstructions.
    The green arrows indicate the recovered assembly motions.
    The rightmost column presents novel-view renderings of the fully assembled objects.
    }
    \label{fig:assembly_steps_real}
\end{figure*}

\section{Additional Results of 3D Assembly Reconstruction}
\label{sec:visual_assembly_steps}
We provide additional visualizations of progressive 3D part assembly reconstruction on both synthetic and real-world data.
These results illustrate how \textit{AssemblyGS} recovers assembly poses and motions from demonstration videos and progressively reconstructs complete 3D assemblies.

\noindent\textbf{Synthetic Data.}
We present additional assembly reconstruction results on the synthetic Assembly-20 benchmark.
Since the number of assembly steps varies considerably across different objects, we divide the results into two figures: \Cref{fig:assembly_steps_syn_long} presents objects with more assembly steps, while \Cref{fig:assembly_steps_syn_short} shows those with fewer steps.
For each assembly step, we visualize the corresponding demonstration video frames, estimated assembly poses and motions, and intermediate 3D Gaussian reconstruction results.
These examples demonstrate the ability of \textit{AssemblyGS} to recover diverse assembly operations and progressively reconstruct complete 3D objects.

\noindent\textbf{Real-World Data.}
\Cref{fig:assembly_steps_real} presents corresponding assembly reconstruction results on real-world data.
Similarly, we visualize the input demonstration videos alongside the recovered assembly poses, motions, and intermediate 3D Gaussian reconstruction results.
These examples further demonstrate the effectiveness of \textit{AssemblyGS} in recovering assembly processes under real-world imaging conditions and occlusions.


\end{document}